\documentclass[10pt,journal,compsoc]{IEEEtran}
\usepackage{multirow}
\usepackage{empheq}
\usepackage{subfigure}
\usepackage{amssymb}
\usepackage{epsfig,graphicx}
\usepackage{epstopdf}

\DeclareMathOperator*{\argmin}{argmin}

\ifCLASSOPTIONcompsoc
  \usepackage[nocompress]{cite}
\else
  \usepackage{cite}
\fi
\ifCLASSINFOpdf
\else
\fi
\begin{document}
%
\title{DeepBinaryMask: Learning a Binary Mask for Video Compressive Sensing}
%
%
%
%

\author{Michael~Iliadis,~\IEEEmembership{Member,~IEEE,}
        Leonidas~Spinoulas,~\IEEEmembership{Member,~IEEE,}
        and~Aggelos~K.~Katsaggelos,~\IEEEmembership{Fellow,~IEEE}
\IEEEcompsocitemizethanks{\IEEEcompsocthanksitem Michael Iliadis, Leonidas Spinoulas and Aggelos K. Katsaggelos are with the Department
of Electrical Engineering and Computer Science, Northwestern University, Evanston, IL 60201, USA
\protect\\
E-mail: miliad@u.northwestern.edu, leonisp@u.northwestern.edu, aggk@eecs.northwestern.edu
}
}

\IEEEtitleabstractindextext{%
\begin{abstract}
In this paper, we propose a novel encoder-decoder neural network model referred to as DeepBinaryMask for video compressive sensing. In video compressive sensing one frame is acquired using a set of coded masks (sensing matrix) from which a number of video frames is reconstructed, equal to the number of coded masks. The proposed framework is an end-to-end model where the sensing matrix is trained along with the video reconstruction. The encoder learns the binary elements of the sensing matrix and the decoder is trained to recover the unknown video sequence. The reconstruction performance is found to improve when using the trained sensing mask from the network as compared to other mask designs such as random, across a wide variety of compressive sensing reconstruction algorithms. Finally, our analysis and discussion offers insights into understanding the characteristics of the trained mask designs that lead to the improved reconstruction quality. 
\end{abstract}

}

\maketitle

\IEEEdisplaynontitleabstractindextext

%
\IEEEpeerreviewmaketitle

\IEEEraisesectionheading{\section{Introduction}\label{sec:introduction}}

%
%
%
%
\IEEEPARstart{I}{N} signal processing, Compressive Sensing (CS) is a popular problem which has been incorporated in various applications~\cite{Donoho2006,Candes2006}. In principle, CS theory suggests that a signal can be perfectly reconstructed using a small number of random incoherent linear projections by finding solutions to underdetermined linear systems. The underdetermined linear system in CS is defined by,
\begin{equation}
{\bf y} = \Phi {\bf x},
\label{eq:measurementModel}
\end{equation}
where $\Phi$ is the $M_{f} \times N_{f}$ measurement or sensing matrix with $M_{f} \ll N_{f}$. We denote the vectorized versions of the unknown signal and compressive measurements as ${\bf x}: N_{f}\times 1$ and ${\bf y}: M_{f} \times 1$, respectively. Thus, having more unknowns than equations, to guarantee a single solution in system \eqref{eq:measurementModel} sparsity on the signal is enforced. Many signals, such as natural images, are sparse in well-known bases (e.g., Wavelet). Therefore, most reconstruction approaches employ a regularization term $F(\cdot)$ which promotes sparsity of the unknown signal ${\bf x}$ on some chosen transform domain. Thus, the following minimization problem is sought after,
\begin{equation}
\label{eq:l1-min}
\begin{aligned}
& {\hat {\bf a}} = \argmin_{\bf a} F({\bf a}) 
& &\text{s.t.} & &  {\bf y} = \Phi {D} {\bf a}, \\
\end{aligned}
\end{equation}
where ${D}$ is a chosen sparse representation transform resulting in a sparse ${\bf a}$, such that ${\bf x} = {D{\bf a}}$. For example, in the case $F = ||{\bf a}||_0$, the problem in Eq. \eqref{eq:l1-min} is translated to a $\ell_0$ minimization problem, which can be solved with standard numerical methods such as Orthogonal Matching Pursuit (OMP) and Basis Pursuit (BP). 

\begin{figure}[!t]
\centering
\includegraphics{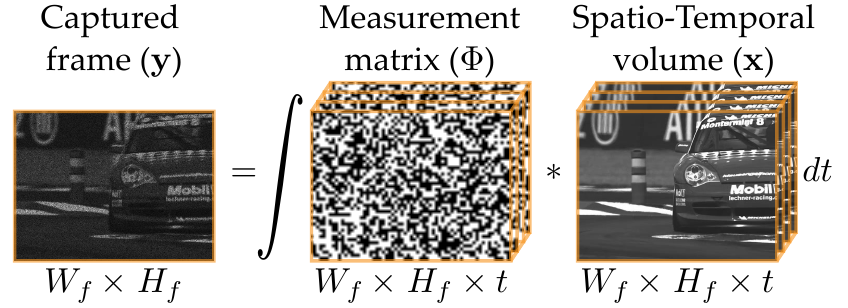}
\caption{Temporal compressive sensing measurement model.}
\label{fig:measurementModel}
\end{figure}

Multiple algorithms have been proposed for reconstructing still
images using CS by solving the problem in \eqref{eq:l1-min}. The problem of video compressive sensing (VCS) refers to the recovery of an unknown spatio-temporal volume from the limited compressive measurements. There are two different approaches in VCS, namely spatial and temporal. Spatial VCS architectures perform spatial multiplexing per measurement based on the well-known single-pixel-camera~\cite{Duarte2008} and enable video recovery by expediting the capturing process~\cite{Sankaranarayanan2012, Wang2015, Chen2015}. In temporal VCS, multiplexing
occurs along the time dimension. Figure~\ref{fig:measurementModel} demonstrates this process, where a spatio-temporal signal of size $W_{f}\times H_{f} \times t = N_{f}$ is modulated by $t$ binary random masks during the exposure time of a single capture and produces a coded frame of size $W_{f} \times H_{f} = M_{f}$. The acquisition model in \eqref{eq:measurementModel} applies to the temporal VCS case as well. However, the construction of $\Phi$ is different in this case. In particular, it is sparse and is given by,
\begin{equation}
\label{eq:vcs}
\Phi = \left[ diag({ \phi}_{1}), \dots, diag({ \phi}_{t})\right] : M_{f} \times N_{f},
\end{equation}
where each vectorized sampling mask is expressed as ${ \phi}_{1}, \dots, { \phi}_{t}$ and $diag(\cdot)$ creates a diagonal matrix from its vector argument. It is noted here that the spatio-temporal volume is lexicographically ordered into the vector ${\bf x}$ by considering first the spatial and then the temporal dimensions. 

Performance guarantees for sparse reconstruction methods, i.e., OMP, indicate that matrix $\Phi$ must be an incoherent unit norm tight frame~\cite{Tsiligianni2015}. Incoherence is a property that characterizes the degree of similarity between the columns of $\Phi$ (or $\Phi{D}$). Therefore, the choice of matrix $\Phi$ is crucial for the reconstructed image and video quality irrespectively of the choice of $F(\cdot)$. For signals that can be represented sparsely in some basis, various popular matrices in the literature are known to perform particularly well (e.g., Gaussian). However, in VCS the design of $\Phi$ as part of the acquisition hardware (e.g., camera) introduces certain limitations. For practical implementations, binary random matrices (e.g., Bernoulli) are better suited while they perform favorably to Gaussian random matrices~\cite{Baraniuk2008}.

The problem of optimizing the ${\Phi}$ matrix has been analyzed by several researchers~\cite{Elad2007,Tsiligianni2014,Tsiligianni2015,Xu2010}. Unfortunately, optimization approaches typically rely on minimizing the coherence between the sampling matrix $\Phi$ and the sparsifying basis ($\Phi{D}$), which mostly applies to spatial compressive sensing where dense matrices are used. Instead, the masks used for temporal VCS systems, as the one described herein, result in a sparse binary matrix with entries across diagonals, as presented by Eq. \eqref{eq:vcs}, and therefore existing results are not applicable. 

In this work, we optimize the sensing matrix $\Phi$ for temporal VCS and transform it into a form that is more suitable for reconstruction using deep neural networks. The proposed neural network architecture, which we refer to as {\em DeepBinaryMask}, consists of two components that act as a pair of an encoder and a decoder. The encoder maps a video block to compressive measurements by learning binary weights (which correspond to the entries of the measurement matrix). The decoder maps the measurements back to a video block utilizing real-valued weights. Both networks are trained jointly. We show that the mask trained from data using neural networks provides significantly improved recovery performance as compared to a non-trained sensing mask.

\subsection{Contributions}
\begin{itemize}
\item {\bf Learning binary weights and reconstruction simultaneously}: We propose a novel encoder-decoder neural network for temporal VCS in which the encoder learns binary weights that form the sensing mask and the decoder learns to reconstruct the video sequence given the encoded measurements.
\item {\bf Learning a general mask}: We show that the reconstruction performance is improved when using the optimized trained mask over a random one. Performance improvements are reported not only when the reconstruction method is the neural network decoder but also when other popular reconstruction methods are employed (e.g., based on $\ell_1$ optimization).
\item {\bf Mask analysis}: We present a reconstruction performance analysis of the trained sensing mask/matrix for different mask initializations (e.g., initial number of nonzero elements).
\end{itemize}


\section{Motivation and Related Work}
\label{sec:motivation}

Recent advances in Deep Neural Networks (DNNs)~\cite{LeCun2015} have demonstrated state-of-the-art performance in several computer vision and image processing tasks, such as image recognition~\cite{He2015} and object detection~\cite{Girshick2015}. In this section we briefly discuss previous works in designing optimal masks for VCS and then we survey recent studies in image recovery problems using DNNs. Finally, we describe advances in DNNs utilizing binary weights, a key ingredient of our proposed method. \newline \newline {\bf Designing optimal masks.} Most of the previously proposed optimized mask patterns for temporal VCS rely on some heuristic constraints and trial-and-error patterns. A thresholded Gaussian matrix was employed in~\cite{Reddy2011}, ~\cite{Llull2013b} and \cite{Gao2014} as it was assumed that it results in a sensing matrix that most closely resembles a dense Gaussian matrix. A normalized mask such that the total amount of light collected at each pixel is constrained to be constant was proposed by~\cite{Liu2014}. It was found in~\cite{Koller2015} that these normalized patterns produce improved reconstruction performance. In~\cite{Koller2015} a hybrid normalized and Gaussian thresholded mask was utilized which was found to outperform the masks proposed in~\cite{Liu2014} and~\cite{Llull2013b}.

Differently from these works, our proposed approach does not impose any mask constraints but instead generates mask patterns learnt from the training data. To the best of our knowledge this is the first study that investigates the construction of an optimized binary CS mask through DNNs. \newline \newline {\bf DNNs for image recovery.} The capabilities of deep architectures have been investigated in image recovery problems such as deconvolution~\cite{Schuler2013,Sun2015,Xu2014}, denoising~\cite{Burger2012,Agostinelli2013,Vincent2010}, inpainting~\cite{Xie2012}, and super-resolution~\cite{Cui2014,Dong2016,Ren2015,Huang2015, Kappeler2016}. Deep architectures have also been proposed for CS of still images. In~\cite{Mousavi2015}, stacked denoising auto-encoders (SDAs) were employed to learn a mapping between the CS measurements and image blocks. A similar approach was also utilized in~\cite{Kulkarni2016} but instead of SDAs, convolutional neural networks (CNNs) were used. 

\begin{figure*}[!t]
\centering
\includegraphics{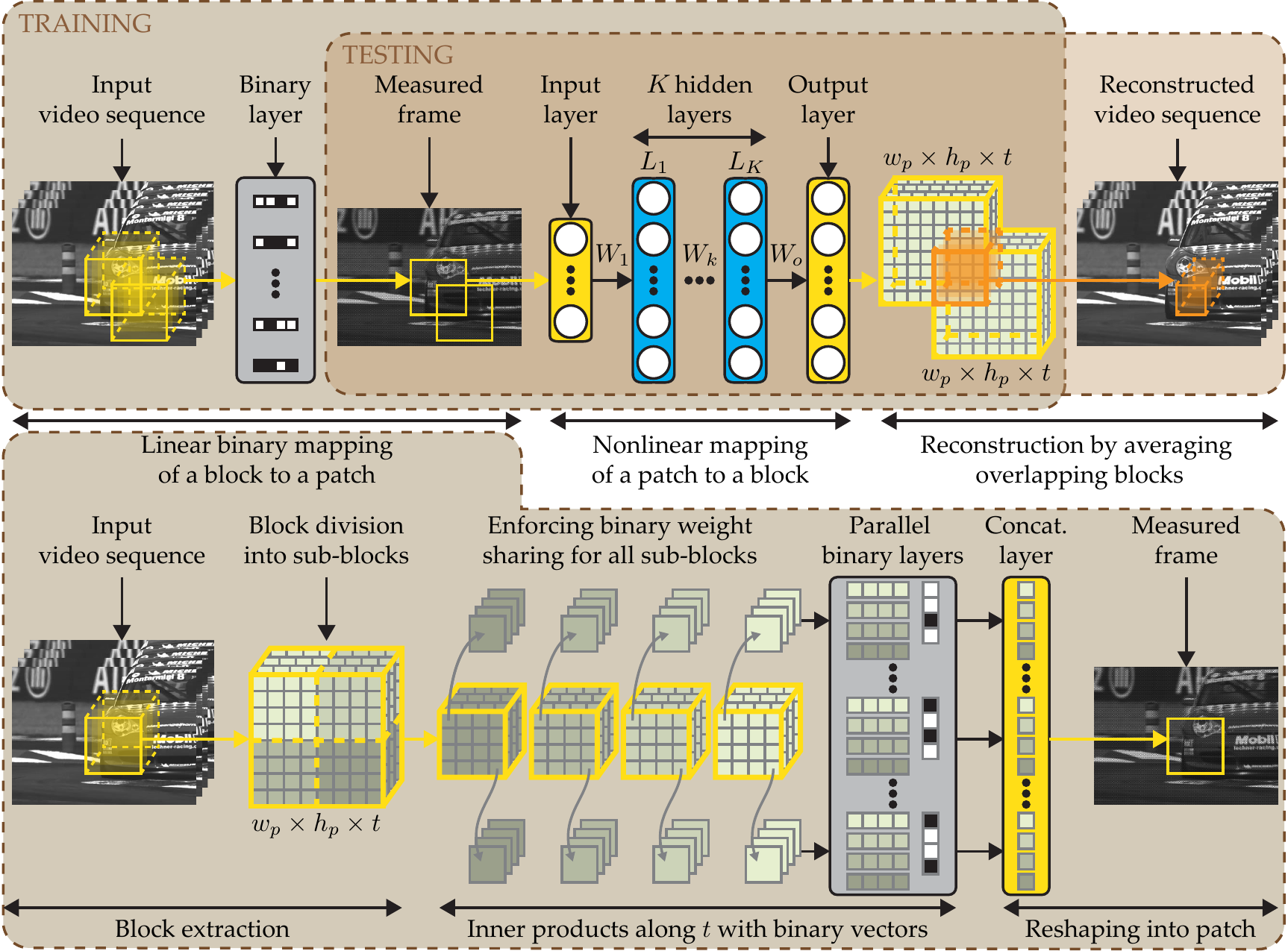}
\caption{Illustration of the proposed encoder-decoder neural network for video compressive sensing. The bottom part demonstrates the {\em encoder} network that is responsible for learning the binary mask and outputs CS measurements. The upper part, labeled as ``TESTING" illustrates the {\em decoder} network which takes as input CS measurements and outputs a video sequence.}
\label{fig:network_pami}
\end{figure*}

A closely related study is our previous work in~\cite{Iliadis2016} which focuses on learning to map directly temporal VCS measurements to video frames using deep fully-connected networks when the measurement matrix is fixed. We showed that the deep learning framework enables the recovery of video frames from temporal compressive measurements in a few seconds at significantly improved reconstruction quality compared to different, optimization based, schemes. \newline \newline {\bf Binary neural networks.} Recently, several approaches have been proposed on the development of neural networks with binary weights~\cite{Courbariaux2016,Courbariaux2015,Lin2015,Rastegari2016} for image recognition applications. The main objective of such an approach is to simplify computations in neural networks, thus making them more efficient while requiring reduced storage. Efficiency is achieved by approximating the standard real-valued DNNs with binary weights. In BinaryConnect~\cite{Courbariaux2015} the authors proposed to binarize the weights for all layers during the forward and backward propagations while keeping the real-valued weights during the parameter update. The real-valued updates were found to be necessary for the application of stochastic gradient descent (SGD). Performance with various classification tasks demonstrated that binary neural networks compare favorably with real-valued weight networks. In~\cite{Rastegari2016}, the authors introduced a weight binarization scheme where both a binary filter and a scaling factor are estimated. Such scheme was proven more effective compared to the BinaryConnect.

Motivated by the success of DNNs in CS reconstruction and binary DNNs in classification, we investigate in this paper the problem of learning an optimized binary sensing matrix using DNNs for temporal VCS.

The work presented in this paper is different from the studies in image recovery using DNNs and from the binary neural networks. First, this work is different from our work in~\cite{Iliadis2016} since our focus is on learning an optimized sensing mask along with the video reconstruction. In~\cite{Iliadis2016} the scope was to recover video frames directly from the temporal measurements (e.g., the mask is pre-defined). Furthermore, our objective in this paper is to learn binary masks that will encode video frames on VCS cameras for video reconstruction which is different from that in binary neural network studies, which is efficiency for image recognition problems.

\section{DeepBinaryMask}
\label{sec:binarymask}

In this work, we propose a novel neural network architecture that learns to {\em encode} a three dimensional (3D) video block to compressive two-dimensional (2D) measurements by learning the binary weights of $\Phi$ and to {\em decode} the measurements back to a video block, as illustrated in Figure~\ref{fig:network_pami}. Let us now describe in detail the encoder and decoder. 

\subsection{Encoder}
\label{subsec:encoder}

In order for our learning approach to be practical, reconstruction has to be performed on 3D video blocks~\cite{Kulkarni2016,Iliadis2016}. Thus, each video block must be sampled with a block-based measurement matrix which should be the same for all blocks. Furthermore, such a measurement matrix should be realizable in hardware. We follow the pattern in~\cite{Iliadis2016} and we consider a $\Phi$ which consists of repeated identical building blocks of size $w_{p} \times h_{p} \times t = N_{p}$ corresponding to the matrix $\Phi_p$ of size $M_p \times N_p$, where $M_p = w_p \times h_p$. In other words $\Phi_p$ has the structure shown in Eq. \eqref{eq:vcs}, in which $M_{f}$ and $N_{f}$ have been respectively replaced by $M_{p}$ and $N_{p}$. An implementation of such a matrix on existing systems employing Digital Micromirror Devices (DMDs), Spatial Light Modulators (SLMs) or Liquid Crystal on Silicon (LCoS)~\cite{Chen2015,Gao2014,Liu2014,Reddy2011,Wang2015} can be easily performed. At the same time, a repeated mask can be printed and shifted appropriately to produce the same effect in systems utilizing translating masks~\cite{Koller2015,Llull2013b}.

Let us consider a set of $N$ training 3D video blocks, each of size $w_{p} \times h_{p} \times t$. They are lexicographically ordered by considering first the spatial and then the temporal dimensions to form vectors ${\bf x}_i$, each of size $N_{p} \times 1$. The encoder is defined as the mapping $g(\cdot)$ that transforms each ${\bf x}_i$ to a measurement ${\bf y}_i$ of size $M_{p} \times 1$, which represents the lexicographically ordered $w_p \times h_p$ image patch, followed by a non-linearity given as,
\begin{equation}
{\bf y}_i = g({\bf x}_i;\theta_e) = \sigma_e(\Phi_{p}{\bf x}_i),
\label{eq:matrixMapping}
\end{equation} 
where $\theta_e = \{\Phi_p\}$ is the parameter set and function $\sigma_e(\cdot)$ represents the non-linearity. We use the subscript ``$e$'' to denote quantities pertaining to the encoder, in order to distinguish them from the decoder quantities to be introduced later.

\begin{figure}[!t]
\centering
\includegraphics{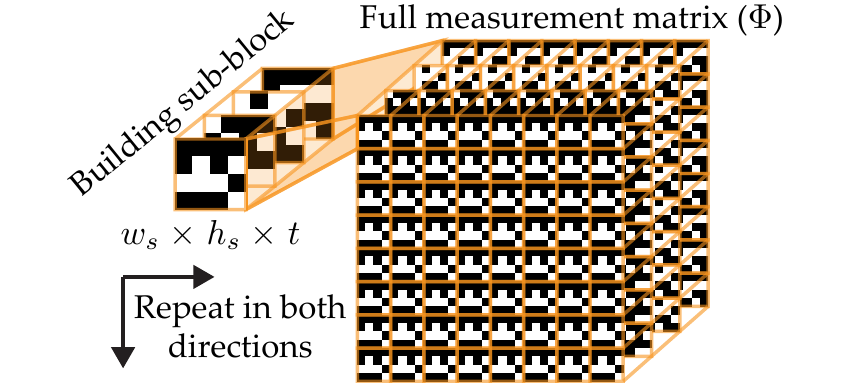}
\caption{Construction of the full measurement matrix by repeating a three dimensional random binary array (building sub-block) in the horizontal and vertical directions.}
\label{fig:maskConstruction}
\end{figure}

The formulation in \eqref{eq:matrixMapping} would have been straightforward to handle if matrix ${\Phi_p}$ were dense and consisting of real-values. However, as mentioned earlier, in the case of temporal VCS, matrix $\Phi$ is binary (due to implementation considerations) and sparse following the structure defined in \eqref{eq:vcs}. For ease of presentation let us now also define a matrix $B$ of size $t\times M_p$ containing the binary weights as,
\begin{equation}
\label{eq:Bmatrix}
B = \left[ {\bf b}_1, \dots {\bf b}_{M_p}\right] =
\left[\begin{array}{ccc}
b_{1,1} & \dots & b_{1,M_p} \\
\vdots & \ddots & \vdots \\
b_{t,1} & \dots & b_{t,M_p}
\end{array} \right].
\end{equation}
It is related to the measurement matrix $\Phi_p$ as,
\begin{equation}
\label{eq:Fmatrix}
\Phi_p = \left[
\begin{array}{ccccccc}
b_{1,1} & {\bf 0} & {\bf 0} &   & b_{t,1} & {\bf 0} & {\bf 0} \\
{\bf 0} & \ddots & {\bf 0} & \cdots & {\bf 0} & \ddots & {\bf 0} \\
{\bf 0} & {\bf 0} & b_{1,M_p} &  & {\bf 0} & {\bf 0} & b_{t,{M_p}}
\end{array} \right].
\end{equation}
In order to realize such a structure in a neural network and be able to train it we transform the encoder into a network that involves the following steps:
\begin{enumerate}
\item The first step consists of $M_{p}$ binary parallel layers. To describe this step we need to introduce a new column ($t \times 1$) vector ${\bf x}_{i,{j}}$, which consists of all the temporal elements at a given spatial location $j$, that is,
\begin{equation}
     {\bf x}_{i,j}=\begin{bmatrix}
         {{\bf x}_{i}(j)} \\
         {{\bf x}_{i}(M_p+j)} \\
         {{\bf x}_{i}(2M_p+j)} \\
         {\vdots} \\
         {{\bf x}_{i}\Big((t-1)M_p+j\Big)}
        \end{bmatrix},
  \end{equation}
where ${{\bf x}_{i}}(j)$ denotes the $j$-th element of vector ${\bf x}_i$. Then in parallel the following inner products are computed,
\begin{equation} 
 e({\bf x}_{i,{j}}) = {\bf b}_{j}^T{\bf x}_{i,j}, \;\;\;  \mbox{ for }  j=1,...,{M_p}. 
 \end{equation}
\item The second step consists of a concatenation layer which concatenates the outputs of the parallel layers in order to construct a single measurement vector that is,
\begin{equation}
{\bf y}_i = g({\bf x}_i;\theta_e) = concat\Big(e({\bf x}_{i,{1}}), \ldots, e({\bf x}_{i,{M_p}})\Big),
\label{eq:concat}
\end{equation}
with a parameter set $\theta_e = \{{\bf b}_1, \dots ,{\bf b}_{M_p}\}$, as defined by Eqs.~\eqref{eq:Bmatrix} and \eqref{eq:Fmatrix}. Note, that a non-linearity such as the rectified linear unit (ReLU)~\cite{Nair2010} defined as, $\sigma(z) = \max(0,z)$, is implicitly applied here after the concatenation since the output is always positive. This is due to the fact that the weights are binary with values $0$ and $1$ and the video inputs have non-negative values.    
\end{enumerate}
The above two steps follow the model presented in Figure~\ref{fig:measurementModel} but translated to a neural network, where the set $\theta_e$ consists of the elements of the trained projection matrix. The two steps of the encoder are illustrated at the bottom part of Figure~\ref{fig:network_pami}. Note that the figure refers to the encoding of overlapping blocks, as we describe next. \newline \newline {\bf Overlapping blocks and weight sharing.} The $t \times M_p$ binary weight matrix $B$ we have considered so far corresponds to non-overlapping video blocks. In order to realize overlapping blocks which usually aid in improving reconstruction quality we can utilize repeating blocks of dimensions $\frac{w_p}{2} \times \frac{h_p}{2} \times t$, which we call sub-blocks as shown in Figure~\ref{fig:network_pami}. Thus, for the final trained matrix $\Phi_p$ each $\frac{w_p}{2} \times \frac{h_p}{2} \times t$ sub-block is the same allowing reconstruction of overlapping blocks of size $w_p \times h_p \times t$ with spatial overlap of $\frac{w_p}{2} \times \frac{h_p}{2} = w_s \times h_s$, as presented in Figure~\ref{fig:maskConstruction}. In such a case the parameter set $\theta_e$ is also different. Instead of learning $M_p$ binary weight vectors we learn $M_p/4$, where each weight vector is {\em shared} four times for each of the corresponding pixel positions of the input. For example, in the case when $w_p \times h_p = 8 \times 8$ there will be four identical $4 \times 4 \times t$ sub-block projection matrices. Notice in Figure~\ref{fig:network_pami} that the values of the input block at the corresponding pixel locations at each of the sub-blocks are multiplied by the same binary vector. Thus, for this example, we only need to estimate $16$ binary weight vectors and each one is shared by four different inputs. For instance, in order to calculate $e({\bf x}_{i,{1}})$, $e({\bf x}_{i,{5}})$, $e({\bf x}_{i,{33}})$, $e({\bf x}_{i,{37}})$ the weight vector ${\bf b}_1$ is used. \newline \newline {\bf Binary weights.} Let us now proceed to describe how to estimate the binary weights. We follow the BinaryConnect method~\cite{Courbariaux2015} to constrain the weights of the encoder to be equal to either $0$ or $1$ during propagation. The binarization scheme to transform the real-valued weights to binary values is based on the sign function, that is,
\begin{equation}
b_b = \begin{cases}
1 &\text{if $b_r \geq 0$,}\\
0 &\text{otherwise},
\end{cases}
\label{eq:signf}
\end{equation}
where $b_b$ and $b_r$ are the binarized and real-valued weights of $B$, respectively. Following the training process in~\cite{Courbariaux2015} we binarize the weights of the encoder only during the forward and backward propagations. The update of the parameter set ${\theta_e}$ is performed using the real-valued weights. As explained in~\cite{Courbariaux2015}, keeping the real-valued weights during the updates is necessary for training the networks using SGD. In addition, we enforced the real-valued weights to lie within the $[-1,1]$ interval at each training iteration. The weight clipping was chosen since otherwise the weights may become infinitely large having no impact during binarization. \newline \newline {\bf Weight initialization.} The network weight initialization of the encoder corresponds in our case to the mask initialization. Typically, in VCS the mask is generated randomly. Similarly here, we start with a randomly generated mask by a Bernoulli distribution. However, since real-valued weights are also required by the network to perform their updates we consider the following initialization scheme,
\begin{equation}
\begin{aligned} 
&b_b \sim \operatorname{Bern} \left({p}\right), \\
&b_r = \begin{cases}
\sim \operatorname{Unif} \left({0,1/\sqrt{t}}\right) &\text{if $b_b = 1$,}\\
\sim \operatorname{Unif} \left({-1/\sqrt{t},0}\right) &\text{otherwise},
\end{cases}
\end{aligned} 
\label{eq:initialization}
\end{equation}
where $\operatorname{Bern}(\cdot)$ and $\operatorname{Unif}(\cdot)$ denote the Bernoulli and Uniform distributions, respectively, $p$ is the probability of the weight to be initialized with $1$ and notation $\left(\cdot,\cdot \right)$ refers to the lower and upper bounds for the values of the distribution. The bounds of the Uniform distribution follow the scheme introduced in~\cite{Xavier2010}. The initialization scheme proposed above allows us to fully understand the benefits of learning as compared to non-learning the mask along with reconstructing the video. This is due to the fact that in the case of choosing the non-learning mode we keep the initial $b_b$ weights, drawn from the Bernoulli distribution, fixed.

\subsection{Decoder}
\label{subsec:decoder}

The resulting hidden measurement ${\bf y}_i$ produced by the encoder is then mapped back to a reconstructed $N_p \times 1$ vector through the decoder $f({\bf y}_i;\theta)$, which when unstacked results in the $w_p \times h_p \times t$ dimensional video block, as illustrated in the upper part of Figure~\ref{fig:network_pami}. Thus, the decoder of the proposed method is another network which is trained to reconstruct the video output sequence given ${\bf y}_i$. We consider a Multi-Layer Perceptron (MLP) architecture to learn a nonlinear function $f(\cdot)$ that maps a measured frame patch ${\bf y}_{i}$ via several hidden layers to a video block ${\bf x}_{i}$ as in~\cite{Iliadis2016}.

The output of the $k^{th}$ hidden layer $L_k$, $k = 1, \dots, K$ is defined as,
\begin{equation}
h_k({\bf y}_i) = \sigma_d({ W_{k} h_{k-1}({\bf y}_i) + {\bf c}}_{k}),\quad \textrm{with } h_0({\bf y}_i) = {\bf y}_i,
\end{equation}
where $W_{k}$ is the output weight matrix, and ${\bf c}_k \in \mathbb{R}^{N_p}$ the bias vector. $W_1 \in \mathbb{R}^{N_p \times M_p}$ connects ${\bf y}_i$, the output of the encoder, to the first hidden layer of the decoder, while for the remaining hidden layers, $\left\{{W}_2,\dots,W_K\right\}\in \mathbb{R}^{N_p \times N_p}$. The last hidden layer is connected to the output layer via ${\bf c}_{o} \in \mathbb{R}^{N_p}$ and $W_o \in \mathbb{R}^{N_p \times N_p} $ without nonlinearity. The non-linear function $\sigma_d(\cdot)$ is the ReLU and the weights of each layer are initialized to random values uniformly distributed in $(-1/\sqrt{N_p},1/\sqrt{N_p})$~\cite{Xavier2010}.

There are several reasons why the MLP architecture for the decoder is a reasonable choice for the temporal video compressive sensing problem which have been explained in~\cite{Iliadis2016}. Since our focus in this work is to primarily investigate and compare the performance of the trained versus the non-trained sensing matrix we adopt the decoder design in~\cite{Iliadis2016}.

\subsection{Training the encoder-decoder network}

The two components of the proposed MLP encoder-decoder are jointly trained by learning all the weights and biases of the model. Using spatial overlap $\frac{w_p}{2} \times \frac{h_p}{2}$ the set of all parameters is denoted by $\theta = \left\{{\bf b}_1,\dots{\bf b}_{M_p/4}; W_1,\dots W_K;W_o; {\bf c}_1,\dots,{\bf c}_K;{\bf c}_o \right\}$ and is updated by the backpropagation algorithm~\cite{Rumelhart1988} minimizing the quadratic error between the set of the encoded mapped measurements $f({\bf y}_i;\theta)$ and the corresponding video blocks ${\bf x}_i$. The loss function is the Mean Squared Error (MSE) which is given by,
\begin{equation}
L(\theta) = \frac{1}{N}\sum_{i = 1}^N{\left\Vert f({\bf y}_i;\theta) - {\bf x}_i\right\Vert^2_2}. 
\end{equation}
The MSE was used in this work since our goal is to optimize the Peak Signal to Noise Ratio (PSNR) which is directly related to the MSE. \newline \newline {\bf Training procedure.} The overall training procedure can be summarized by the following steps:

\begin{enumerate}
\item Forward propagation is performed by using weights $B$ after binarization in the encoder and real-valued weights $W$ in the decoder. 
\item Then, backpropagation is performed to compute the gradients with respect to layer's activation knowing $B$ and $W$. 
\item Parameter updates are computed using the real-valued weights for both encoder and decoder.
\end{enumerate}

Note that one other difference between our work and~\cite{Courbariaux2015} is that our encoder-decoder neural network does not utilize binary weights in all layers; instead it utilizes binary weights at the encoder and standard real-valued weights at the decoder. \newline \newline {\bf Implementation details.} Our encoder-decoder neural network is trained for $480$ epochs using a mini-batch size of $200$. We used SGD with a momentum set equal to 0.9. We further used $\ell_2$ norm gradient clipping to keep the gradients in a certain range. Gradient clipping is a widely used technique in recurrent neural networks to avoid exploding gradients~\cite{Pascanu2013}. The threshold of gradient clipping was set equal to $0.1$.

One hyper-parameter that was found to affect the performance in our approach is the learning rate. Based on experimentation we chose a starting learning rate for the encoder that was 10 times larger than that for the decoder. This was found to be important as we wanted the weights of the encoder to have their sign changed during the training iterations. In addition, the learning rate was divided by 2 at every 10 epochs in the encoder and by 10 after 400 epochs in the decoder.

All hyper-parameters were selected after cross-validation using a validation test set. \newline \newline {\bf Test inference.} Once the encoder-decoder  neural network is trained we use the trained sensing matrix $B \,\to\,\Phi$ to calculate the compressive measurements ${\bf y}$. Then, given ${\bf y}$ we can use any VCS algorithm (in addition to the decoder network) to reconstruct the video blocks.

\begin{figure*}[!t]
\centering
\includegraphics{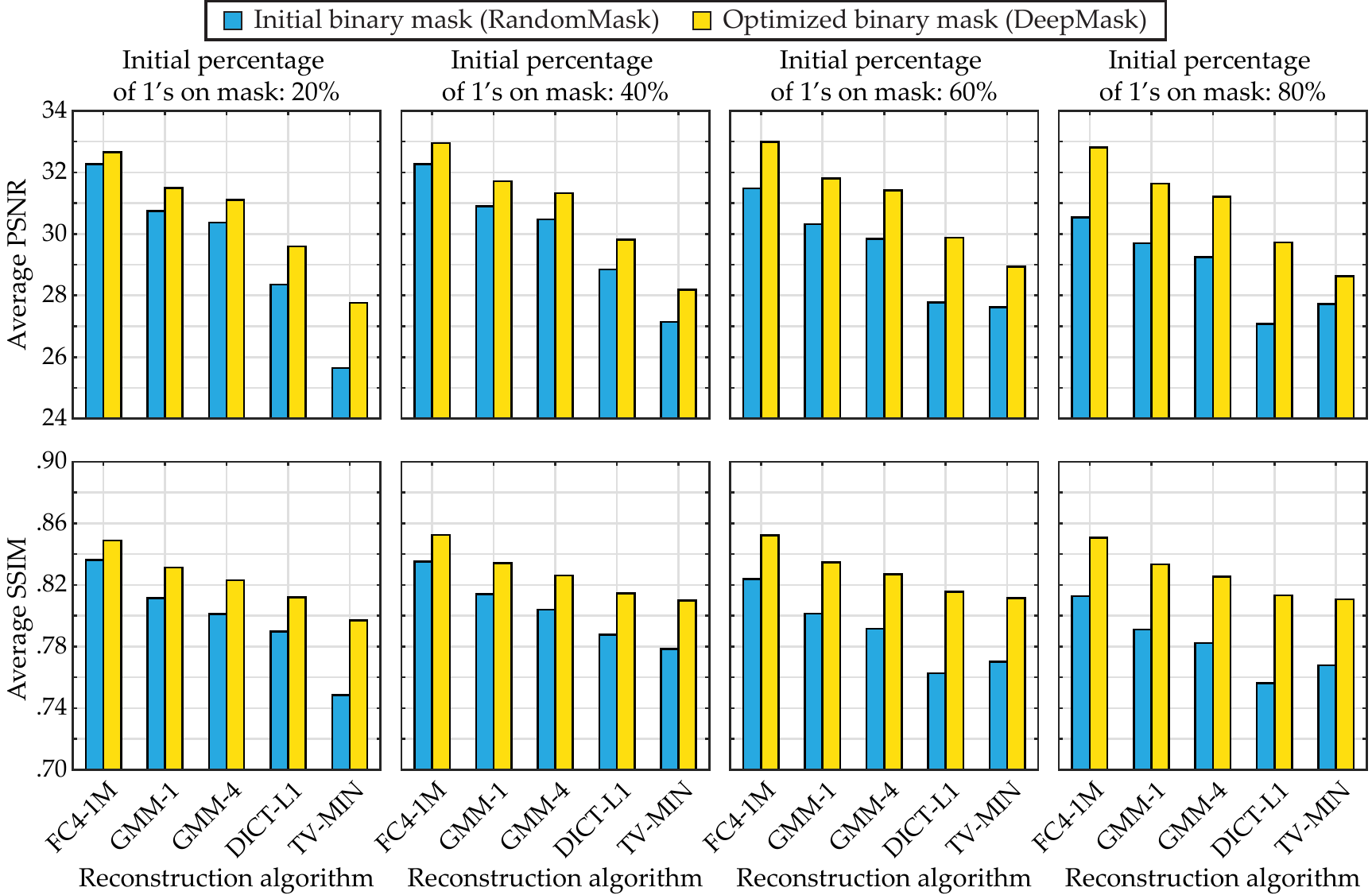}
\caption{Average PSNR and SSIM over all test video sequences for several reconstruction methods using the RandomMasks and DeepMasks. The test set consists of $14$ videos sequences and the reported PSNR and SSIM corresponds to the average values for the reconstruction of the first $32$ frames of each sequence. The PSNR metric is measured in dB while the SSIM is unitless.}
\label{fig:alg_comparison}
\end{figure*}

\section{Experimental Results}
\label{sec:experiments}

In this section we present quantitative and qualitative reconstruction results to demonstrate the effectiveness of the proposed projection mask in temporal VCS. The performance of our trained masks is investigated using various reconstruction algorithms and initial mask parameters. Our analysis offers insights into understanding how the different initial parameters of the mask affect reconstruction performance. The metrics used for reconstruction evaluation were the PSNR and SSIM (Structural SIMilarity). 

\subsection{Training Data Collection and Test set}
\label{subsec:training_data}

In order to train our encoder-decoder architecture we collected a diverse set of training samples using $400$ high-definition videos from Youtube, depicting natural scenes. The video sequences contain more than $10^5$ frames which were converted to grayscale. We randomly extracted $1$ million video blocks of size $w_p \times h_p \times t$ to train our encoder-decoder neural network while keeping the amount of blocks extracted per video proportional to its duration. 

Our test set consists of $14$ video sequences that were used in~\cite{Liu2014} which are provided by the authors. We also included in the test set the ``Basketball'' video sequence used in~\cite{Yang2015}. All test video sequences are unrelated to the training set.

\begin{figure*}[!t]
\centering
\includegraphics{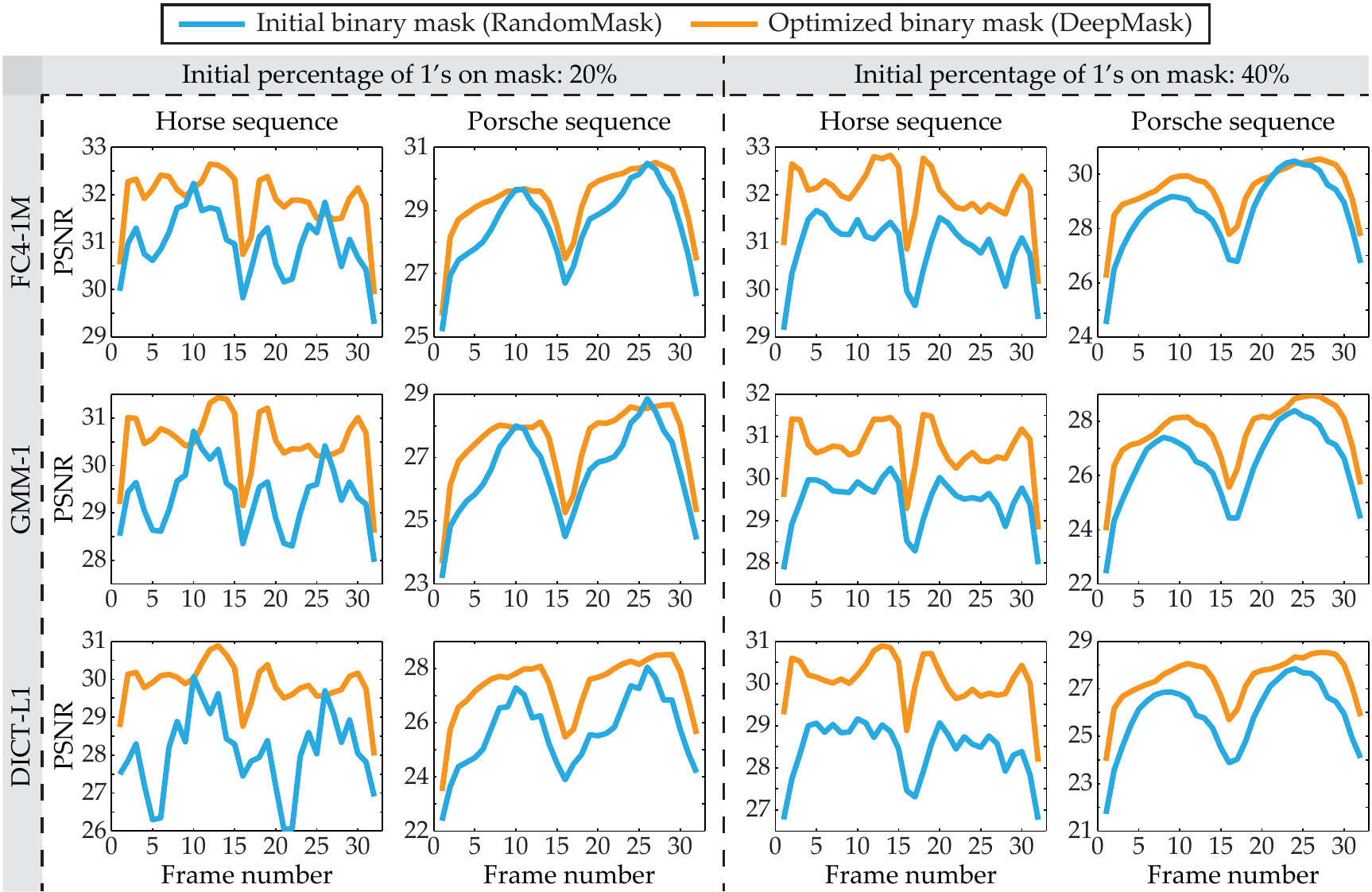}
\caption{PSNR (in dB) comparison for the first $32$ frames of $2$ video sequences among the proposed method FC4-1M and the previous methods GMM-1~\cite{Yang2014} and DICT-L1~\cite{Liu2014}. Notice that the vertical scale changes among the various plots.}
\label{fig:frames_comparison}
\end{figure*}

\begin{figure*}[!t]
\centering
\includegraphics{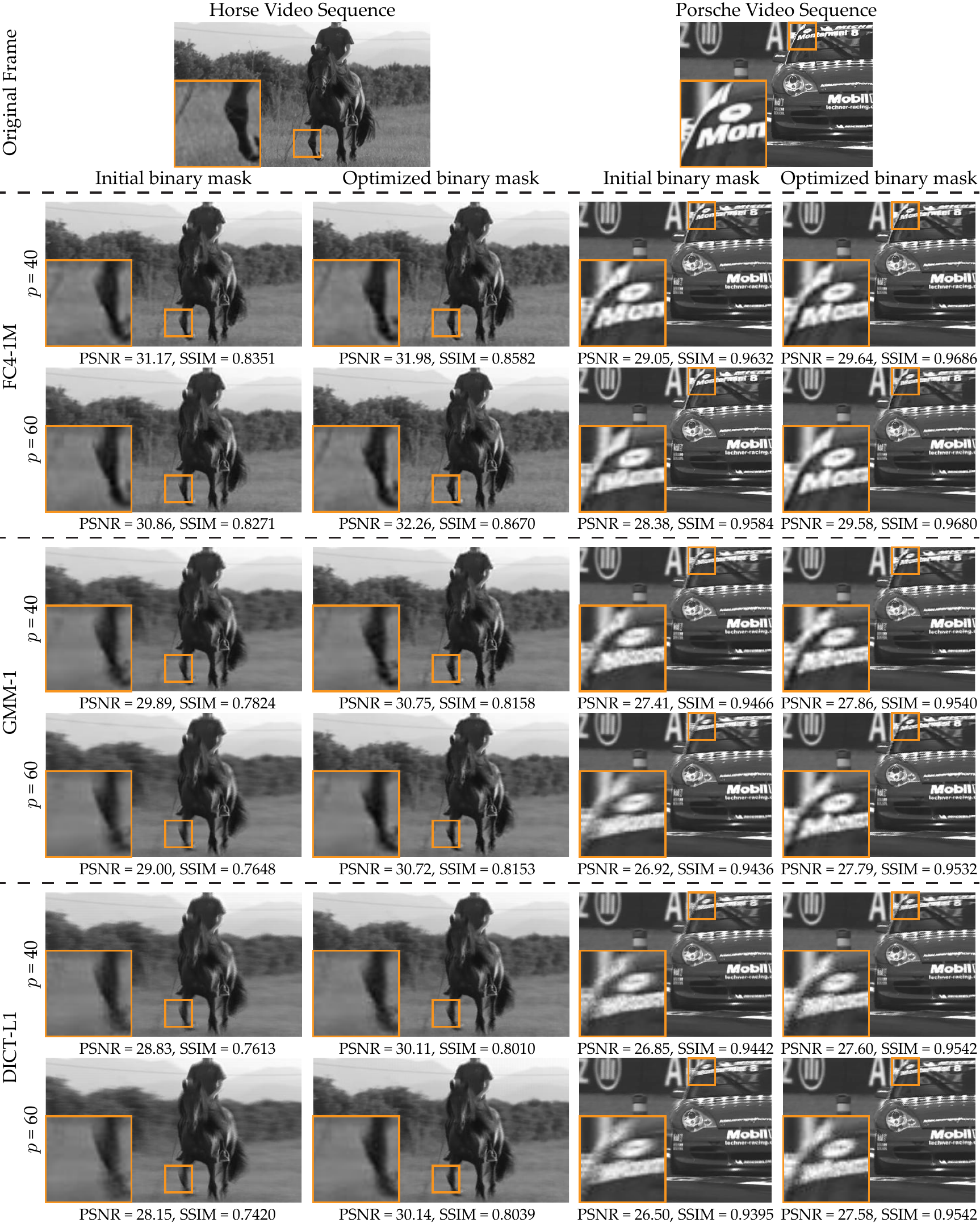}
\caption{Qualitative reconstruction performance. The figure shows reconstruction of a single frame for $2$ test video sequences when using different reconstruction algorithms and different mask initializations. The PSNR metric is measured in dB while the SSIM is unitless.}
\label{fig:framesAndInsets}
\end{figure*}

\subsection{Mask Patterns and Decoding Layers}
\label{subsec:mask_patterns}

Our experimental investigation is motivated by the following two questions: 1) ``How does performance of trained and non-trained masks compare using different reconstruction algorithms?" and 2) ``Does the training procedure result in a unique sensing matrix $\Phi$ irrespectively of the initialization parameters?"  

In order to answer these two questions we simulated noiseless compressive video measurements by realizing four different $\frac{w_p}{2} \times \frac{h_p}{2} \times t$ mask patterns. We denote by ``RandomMask-$p$" the mask that is initialized with $\operatorname{Bern} \left({p}\right)$, as in Eq.~\eqref{eq:initialization}, and is not learnt (that is, the elements of the encoder are fixed). We also denote by ``DeepMask-$p$" the learnt mask trained by our proposed encoder-decoder network described in section~\ref{sec:binarymask} and which is initialized by $\operatorname{Bern} \left({p}\right)$ as in Eq~\eqref{eq:initialization}. Thus, we consider the following four mask patterns:
\begin{itemize}
  \item RandomMask-20 and DeepMask-20, with $p=20\%$.
  \item RandomMask-40 and DeepMask-40, with $p=40\%$.
  \item RandomMask-60 and DeepMask-60, with $p=60\%$.
  \item RandomMask-80 and DeepMask-80, with $p=80\%$. 
\end{itemize}

For the remainder of this paper, we describe the selection of block sizes of $w_{p} \times h_{p} \times t = 8 \times 8 \times 16$, such that $N_{p} = 1024$ and $M_{p} = 64$. Therefore, the compression ratio is $1/16$. In addition, for each of the eight $\Phi$ mask types above, each $\frac{w_p}{2} \times \frac{h_p}{2} \times t = 4 \times 4 \times 16$ block is the same allowing reconstruction for overlapping blocks of size $8 \times 8 \times 16$ with spatial overlap of $4 \times 4$. Note that the same random seed was utilized for all patterns.

Further, we used $K = 4$ hidden layers for the decoder architecture of Figure~\ref{fig:network_pami}. We found out experimentally that for the number of training data used ($1$ million) $4$ layers provided the best performance. A similar observation was reported in~\cite{Iliadis2016} where the addition of extra layers for this number of training data did not lead to performance improvement.

In the following section we present the reconstruction algorithms used to test the eight mask types.

\subsection{Reconstruction Algorithms}

Since our main goal is to compare the performance between a trained sensing mask over a non-trained one in an implementation agnostic to mask patterns, we tested a number of different reconstruction algorithms. Candidate reconstruction algorithms were selected for their utility in solving the underdetermined system in the VCS setting. We evaluated the following optimization algorithms as potential solvers:
\begin{enumerate}
\item {\bf DICT-L1:} In \eqref{eq:measurementModel}, we have assumed that data are noise-free. However, real data are typically noisy and dealing with small dense noise is required. In order to deal with such noise we transform the problem in \eqref{eq:l1-min} into the LASSO (Least-Absolute Shrinkage and Selection Operator) problem for $F = \lambda||{\bf a}||_1$ given as,
\begin{equation}
\label{eq:l1-lasso}
{\hat {\bf a}} = \argmin_{\bf a} \left\lVert{\bf y} - \Phi {D} {\bf a}\right\rVert_2^2 + \lambda||{\bf a}||_1,
\end{equation}
where $\lambda > 0$ is the regularization parameter whose value is related to the noise tolerance. For this problem we chose to use an overcomplete dictionary ${D}$ as a sparsifying basis. The dictionary consists of $20,000$ atoms trained on a subset of $200,000$ video blocks from our training database and reconstruction is performed block-wise on overlapping sets of $7 \times 7$ patches of pixels. For the optimization problem in~\eqref{eq:l1-lasso}, $\lambda$ was set equal to $0.005$.

\item {\bf TV-MIN:} A popular CS reconstruction method utilizes for $F(\cdot)$ the total variation (TV) norm defined as, 
\begin{equation}
\label{eq:tv}
\begin{aligned}
TV({\bf z}) = \sum_{i,j,n}&\Bigg(\Big({z}(i+1,j,n) - {z}(i,j,n)\Big)^2 \\ 
& + \Big({z}(i,j+1,n) - {z}(i,j,n)\Big)^2\Bigg)^{1/2},
\end{aligned}
\end{equation}
where ${\bf z}$ is the stacked version of the 3D array $z(i,j,n)$, where $(i,j,n)$ are respectively the two spatial and one temporal coordinates. Thus, the TV minimization problem is given as,
\begin{equation}
\label{eq:tv-lasso}
{\hat {\bf x}} = \argmin_{\bf x} \left\lVert{\bf y} - \Phi{\bf x}\right\rVert_2^2 + \lambda TV({\bf x}).
\end{equation}
In order to solve \eqref{eq:tv-lasso} we used the two-step iterative shrinkage/thresholding (TwIST) algorithm~\cite{Bioucas-Dias2007} with $\lambda = 0.01$.

\item {\bf GMM-TP:} Another reconstruction algorithm we considered in our experiments is a Gaussian mixture model (GMM)-based algorithm~\cite{Yang2014} learned from Training Patches (TP) referred as GMM-TP. We followed the settings proposed by the authors and used our training data (randomly selecting $20,000$ samples) to train the underlying GMM parameters. In our experiments we refer to this method by GMM-$4$ and GMM-$1$ to denote reconstruction of overlapping blocks with spatial overlap of $4 \times 4$ and $1 \times 1$ pixels, respectively.

\item {\bf FC4-1M:} Finally, another reconstruction method we considered is the decoder neural network introduced in subsection~\ref{subsec:decoder}. The decoder is a $K=4$ MLP trained on $1$ million samples similarly to~\cite{Iliadis2016}. In this case, a collection of overlapping patches of size $8 \times 8$ is extracted by each coded measurement of size $W_f \times H_f$ and subsequently reconstructed into video blocks of size $8 \times 8 \times 16$. Overlapping areas of the recovered video blocks are then averaged to obtain the final video reconstruction results as shown in the upper part of Figure~\ref{fig:network_pami}. The step of the overlapping patches was set to $4 \times 4$ due to the special construction of the utilized measurement matrix, as discussed in subsection~\ref{subsec:encoder}.

\end{enumerate}

For each algorithm, $\lambda$ values were determined based on the best performance among different settings. All code implementations are publicly available provided by the authors while the deep network architectures were implemented in Torch7 \cite{Collobert2011}, a Lua library that allowed us to develop an optimized GPU code.

\subsection{Reconstruction Results}

For each reconstruction algorithm described above, we tested the eight mask types presented in subsection~\ref{subsec:mask_patterns}. \newline \newline {\bf Quantitative and qualitative results.} Figure~\ref{fig:alg_comparison} shows average reconstruction quality for each mask and algorithm combination, using the PSNR and SSIM metrics. The presented metrics refer to average performance for the reconstruction of the first $32$ frames of each test video sequence, using $2$ consecutive captured coded frames for each of the eight masks for every algorithm. First, we note that the DeepMasks perform consistently better compared to the RandomMasks across all reconstruction algorithms and initial percentage of nonzeros. In particular, we observe an improvement around $1$-$2$ dB, in terms of PSNR between the trained and non-trained masks across all initial percentages and algorithms. Furthermore, we observe that the decoder FC4-1M demonstrates the highest PSNR and SSIM values among all algorithms.

Figure~\ref{fig:frames_comparison} compares the PSNR for each of the $32$ frames of $2$ video sequences (``Horse'' and ``Porsche'' sequences) using our FC4-1M algorithm and the previous methods GMM-1~\cite{Yang2014} and DICT-L1~\cite{Liu2014} between the RandomMasks and DeepMasks. The varying PSNR performance across the frames of a $16$ frame block is consistent for all algorithms and is reminiscent of the reconstruction tendency observed in other video CS papers in the literature~\cite{Koller2015,Llull2013b,Yang2015,Yang2014}. Please notice that the scale on the vertical axes of the plot varies.

Finally, Figure~\ref{fig:framesAndInsets} compares the reconstruction quality between the optimized and non-optimized masks for a single frame from the $2$ video sequences under different algorithms and different mask initializations. It is clear that reconstruction quality improves when using the optimized masks, especially for the case when the initial percentage of nonzeros is $p=60$. At the same time, the proposed end-to-end optimization using the deep network (FC4-1M) provides the highest visual quality among the candidate compared algorithms.

\subsection{Training analysis}

\begin{figure}[!t]
\centering
\includegraphics{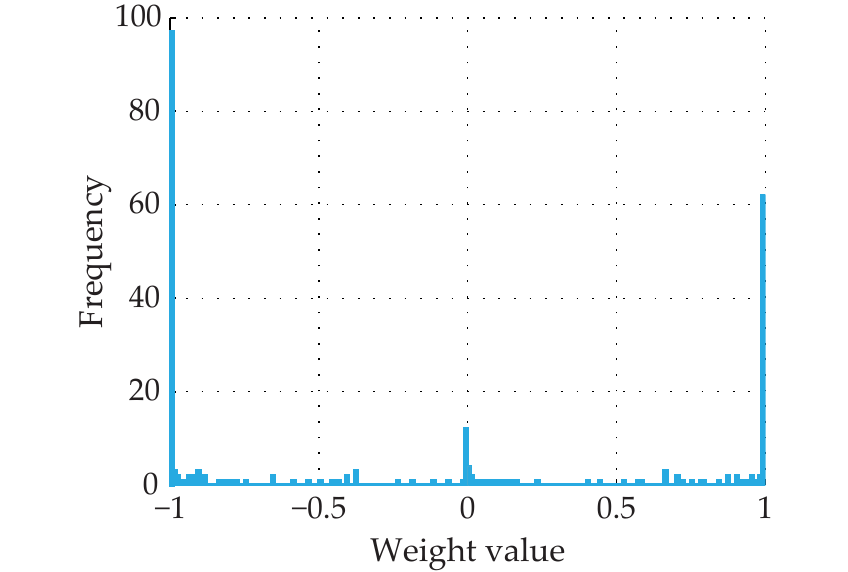}
\caption{Histogram of the real-valued weights produced by the encoder neural network for DeepMask-$40$. We report similar observations with~\cite{Courbariaux2015} as we found out that most of the weights have the tendency to become deterministic ($-1$ and $1$) and reduce the training error while some stay around zero. }
\label{fig:weight_hist}
\end{figure}

\begin{figure}[!t]
\centering
\includegraphics{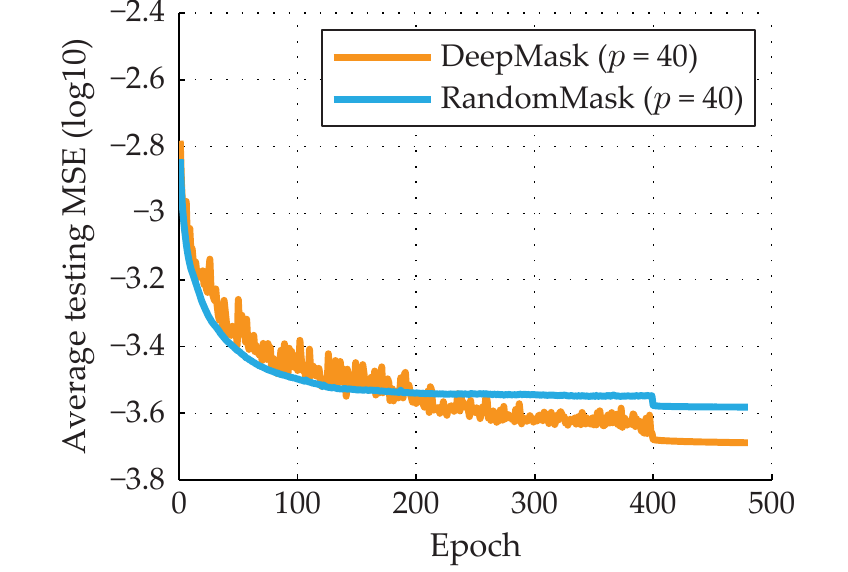}
\caption{Test error curves between the RandomMask-$40$ decoder and DeepMask-$40$ encoder-decoder calculated on a validation test set. The latter provided lower test error upon convergence as is optimized end-to-end.}
\label{fig:test_error}
\end{figure}

\begin{figure}[!t]
\centering
\includegraphics{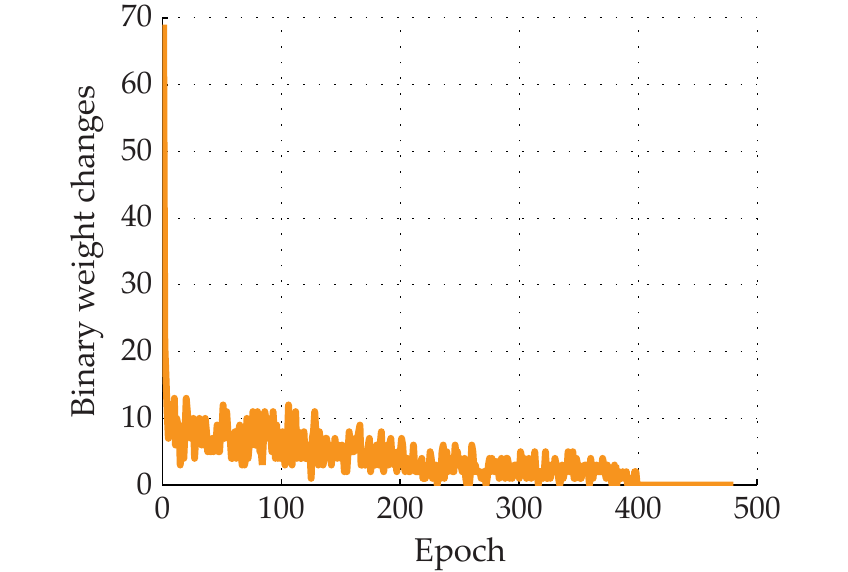}
\caption{Number of binary weight changes per epoch of DeepMask-$40$ encoder-decoder. A large number of weights change in the first few epochs; this number decreases and finally becomes zero in the last few epochs.}
\label{fig:weight_change}
\end{figure}

We start our analysis by examining the real-valued weight histogram of the encoder (DeepMask-$40$) upon convergence in Figure~\ref{fig:weight_hist}. First, we observe that negative values are more frequent than positive ones, which suggests that zero elements of the mask (after binarization) are more important than the nonzero ones. More importantly, we observe that a number of weights are around zero, hesitating between becoming negative or positive, a phenomenon that was also reported in~\cite{Courbariaux2015}. 

Average test MSE per epoch calculated on a validation test set for DeepMask-$40$ and RandomMask-$40$ is shown in Figure~\ref{fig:test_error}. It is shown that the test error curve of the RandomMask-$40$ is smooth while the DeepMask-$40$ is noisy. This is due to the fact that many binary weights switch between $1$ and $0$ frequently, especially during the first epochs of training when the learning rate has high values, thus constantly changing the way the VCS measurements are performed. Furthermore, even at the later stages of training, many real-valued elements of the encoder remain around zero, as observed in Figure~\ref{fig:weight_hist}. Therefore, during the binarization process some of the encoder's binary weights change from zero to one and vice versa even with very small learning rates. However, as the encoder's learning rate becomes really small the curve becomes smooth. A better optimized learning rate decay schedule of the encoder would have probably provided a smoother curve and perhaps a higher performance. We leave this as a task for future work as further investigation into this may be needed. Finally, as showed in the reconstruction results, DeepMask performs consistently better than RandomMask which also explains the lower test MSE produced by the former during training. 

Lastly, in Figure~\ref{fig:weight_change} we show the number of binary weights that change from zero to one and vice versa per epoch. We observe that a large number of weights change in the first few epochs and this number decreases as the number of epochs increases.

\section{Discussion}

Having obtained better reconstruction performance using the DeepMasks across a wide range of reconstruction algorithms our next step is to analyze the masks produced by the networks and highlight a few crucial points. We start our analysis by posing the following question. \newline \newline {\bf Does DeepMask produce a unique sensing matrix $\Phi$?} To answer this question we examine the differences between the produced DeepMasks with respect to their percentage of nonzero elements and to their support. 

\begin{figure}[!t]
\centering
\includegraphics{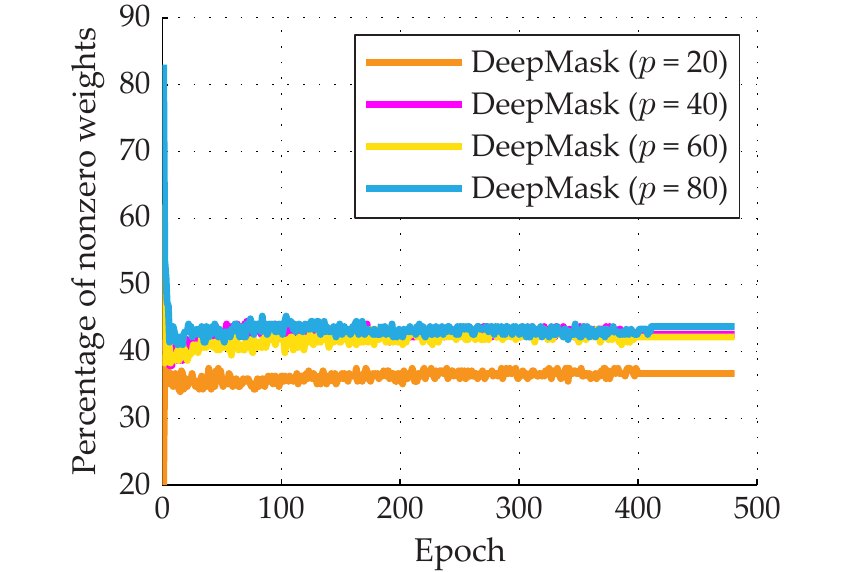}
\caption{Percentage of nonzero binary values for DeepMasks. Irrespectively to the initial nonzero percentage, DeepMasks converge to a point around 40\% nonzeros. }
\label{fig:nonzero_perc}
\end{figure}

First, in Figure~\ref{fig:nonzero_perc} we show the percentage of nonzero binary weights per epoch for the different DeepMasks. This figure allows us to examine uniqueness with respect to the percentage of nonzero elements produced by each DeepMask. We observe that the masks with $p$ equal $40$, $60$ and $80$ converge to a percentage around 40\%. The $p=20\%$ mask though, converges to a bit lower percentage (around 38\% nonzero binary elements). 

\begin{figure}[!t]
\centering
\includegraphics{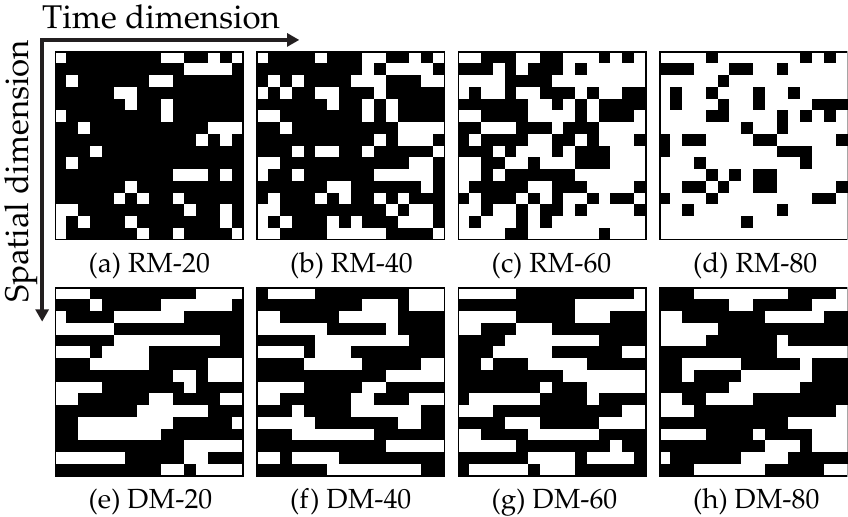}
\caption{The eight mask patterns produced in this work of size $16 \times 16$. The initial RandomMasks (RM) are presented on the top row while on the bottom row we present the optimized DeepMasks (DM).}
\label{fig:masks}
\end{figure}

\begin{figure*}[!t]
\centering
\includegraphics{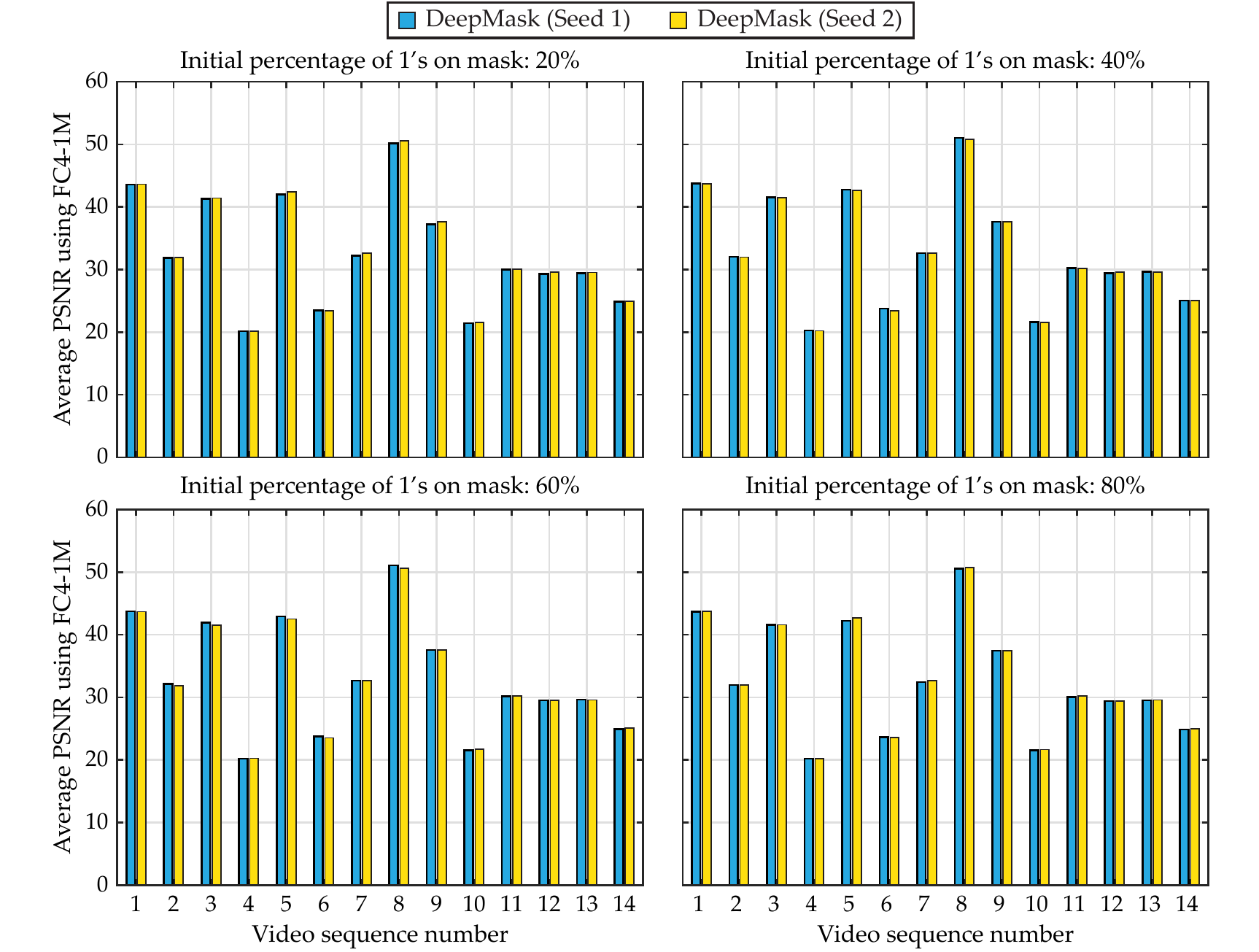}
\caption{Comparison of the reconstruction performance of the proposed encoder-decoder architecture when using the optimized masks trained after initialization with two different seeds. Average PSNR for the reconstruction of the first $32$ frames of each one of the $14$ test video sequences is presented and the values are found to be very similar regardless of the starting binary values of the measurement matrix.}
\label{fig:twoSeeds}
\end{figure*}

Next, Figure~\ref{fig:masks} demonstrates the four mask patterns produced in this work. The first row illustrates the RandomMasks and the second row presents the four DeepMasks produced by the proposed encoder-decoder neural network. All $4 \times 4 \times 16$ masks are reshaped into a $16 \times 16$ matrix for better visualization by lexicography ordering each $4 \times 4$ mask at a given time instance into a $16 \times 1$ vector, which becomes a column of the $16 \times 16$ matrix. That is, the vertical direction denotes pixel location while the horizontal direction denotes time. From the visualization we deduce that although the masks generated by the network converge to the same nonzero percentage (as shown in Figure~\ref{fig:nonzero_perc}), their support is different. The fact that the optimized masks contain a very similar percentage of nonzeros while producing improved reconstruction quality with various different reconstruction algorithms implies that such percentage is the most appropriate one for the task at hand. Similar observations about the ideal percentage of nonzeros for VCS measurement matrices have been made in~\cite{Koller2015}, albeit deduced through heuristic experimentation.

Finally, from the visualization we deduct two important findings: 

\begin{itemize}
\item First, it is apparent that regardless of the initial realization (shown in RandomMasks), the trained DeepMasks produce a similar number of nonzero elements which confirms our findings discussed earlier.
\item Second, an important observation from Figure~\ref{fig:masks} is that DeepMasks are {\em smoother} over time than the RandomMasks. In other words in many rows the binary weights seem to be sequential (or more structured) forming runs of $1$s and $0$s. Again, such finding was heuristically observed in~\cite{Koller2015} and some studies cited therein, further strengthening our findings which are here obtained through a machine learning approach.
\end{itemize}

To summarize, our observations above suggest that an optimized mask design $\Phi$ for temporal VCS incorporates the following two characteristics: 1) {\em smoothness} as explained above and 2) percentage of {\em nonzero elements around 40\%}.

Finally, we wanted to confirm that the results presented herein are not due to a specific selection of the random seed, used to produce the initial random masks. Therefore, we performed the whole training process described in section~\ref{sec:experiments} using a second seed for four new masks and compared the average reconstruction performance using the trained network for all test video sequences. The corresponding results are presented in Figure~\ref{fig:twoSeeds} where it can be observed that the final performance is very similar for both seeds. We do not include the results of the competitive algorithms with this second initialization but observed performance improvements through the new optimized masks similar to the ones presented in Figure~\ref{fig:alg_comparison}.

\section{Conclusions}

In this paper, we proposed a new encoder-decoder neural network architecture for video compressive sensing that is able to learn an optimized binary sensing matrix. We evaluated the proposed model on several video sequences and we documented the superiority of the trained sensing matrices over the random ones both quantitatively and qualitatively. Our qualitative analysis of the trained model shows that the optimized sensing masks converge to a similar number of nonzero elements regardless of their initial parameters and that they exhibit a smoothness property. The proposed architecture has large potential for further analysis. Our next step is to examine the reconstruction performance in real video sequences acquired by a temporal compressive sensing camera.


%

%
%
%
%
%

\ifCLASSOPTIONcaptionsoff
  \newpage
\fi

\bibliographystyle{IEEEtran}
\bibliography{binary_cs}

\end{document}